\documentclass{visuAAL-GoodBrother}
\usepackage[T1]{fontenc}
\usepackage{ucs}
\usepackage{graphicx}
\usepackage{url}
\usepackage[export]{adjustbox}

\emergencystretch=\maxdimen
\begin{document}

\title{Facial Attribute Based Text Guided Face Anonymization}

\author{Mustafa İzzet Muştu, Hazım Kemal Ekenel}
\affiliation{Istanbul Technical University}
\email{\{mustu18, ekenel\}@itu.edu.tr}

\conferenceYear{2024}
\conferenceName{Joint visuAAL-GoodBrother Conference on trustworthy video- and audio-based assistive technologies}
\paperType{COST Action CA19121 - Network on Privacy-Aware Audio- and Video-Based Applications for Active and Assisted Living} 

\maketitle

\begin{abstract}
The increasing prevalence of computer vision applications necessitates handling vast amounts of visual data, often containing personal information. While this technology offers significant benefits, it should not compromise privacy. Data privacy regulations emphasize the need for individual consent for processing personal data, hindering researchers' ability to collect high-quality datasets containing the faces of the individuals. This paper presents a deep learning-based face anonymization pipeline to overcome this challenge. Unlike most of the existing methods, our method leverages recent advancements in diffusion-based inpainting models, eliminating the need for training Generative Adversarial Networks. The pipeline employs a three-stage approach: face detection with RetinaNet, feature extraction with VGG-Face, and realistic face generation using the state-of-the-art BrushNet diffusion model. BrushNet utilizes the entire image, face masks, and text prompts specifying desired facial attributes like age, ethnicity, gender, and expression. This enables the generation of natural-looking images with unrecognizable individuals, facilitating the creation of privacy-compliant datasets for computer vision research.
\end{abstract}


\section*{Introduction}

As computer vision applications become more common, they involve handling massive amounts of visual data, often containing individuals' information. There are many benefits of computer vision technology for the safe driving of autonomous vehicles, home security, and video calls. However, these should not be used at the cost of our privacy. Data privacy is a growing concern, with regulations like the General Data Protection Regulation (GDPR) in the European Union being passed to protect it. As \cite{gdpr} stated, the GDPR affects all processing of personal data including images of individuals and requires the consent of the individual for processing their data. This makes it difficult for computer vision researchers to collect high-quality datasets that include people.

To overcome this concern, anonymization of people appearing in the acquired images becomes a must. While traditional anonymization methods like blurring, pixelization, or cropping faces are common ways to anonymize images, they distort the data so much that these images can not be used for any other purposes later. 
To prevent this problem, recently, realistic anonymization approaches have been proposed (\citet{deidentification}; \cite{le2020anonfaces}; \cite{li2019anonymousnet}; \cite{liu2019a3gan}; \cite{CIAGAN}; \cite{cfanet}). These methods utilize deep learning models to create realistic-looking faces to replace the original ones with them. 

In this work, we present a deep learning-based three-stage face anonymization pipeline. Our pipeline starts with RetinaFace for face detection (\cite{deng2019retinaface}). We employ VGG-Face\footnote{We utilized the DeepFace repository for RetinaFace and VGG-Face (\cite{retina}).} to extract facial attribute information (\cite{vggface}). We incorporate these facial attributes into text and perform anonymization with text guidance by using BrushNet (\cite{brushnet}). In this way, we generate synthetic faces by conserving the age, gender, expression, and ethnicity information from the original image.



\section*{Related Work}

In the current literature, there are two main types of image-generation methods. Most of the methods are derived from GAN which was first proposed by \cite{gans}. The second main approach, the diffusion model, in image generation tasks comes from \cite{diffusion}. For the face anonymization task, the former is the most common method. 




\textbf{Anonymization.} \citet{DeepPrivacy} detect faces using Dual Shot Face Detector (DSFD) and estimate face landmarks using a modified Mask R-CNN (\cite{dsfd}; \cite{maskrcnn}). They use a customized U-Net architecture in the GAN to generate images (\cite{unet}).

\cite{DeepPrivacy2} propose an architecture that can replace faces as well as bodies with synthetic images. They detect and segment faces or whole bodies using three types of detectors. DSFD is used for face detection, Continuous Surface Embeddings (CSE), and Mask-RCNN are used for segmentation. Synthetic data is generated by a style-based GAN (\cite{cse}).

In the Latent Diffusion Face Anonymization (LDFA), \cite{LDFA} use the stable diffusion model from \cite{stabledif}.  The system works in two steps: first, it finds faces with RetinaFace and then replaces them with generated face images from stable diffusion.

\cite{barattin2023attribute} use a pre-trained GAN and proposed directly editing the latent space codes of the synthetic images to ensure a desired distance from the original identity while preserving facial attributes.


\section*{Method}

\begin{figure}[htb]
\centerline{\includegraphics[width=0.6\paperwidth]{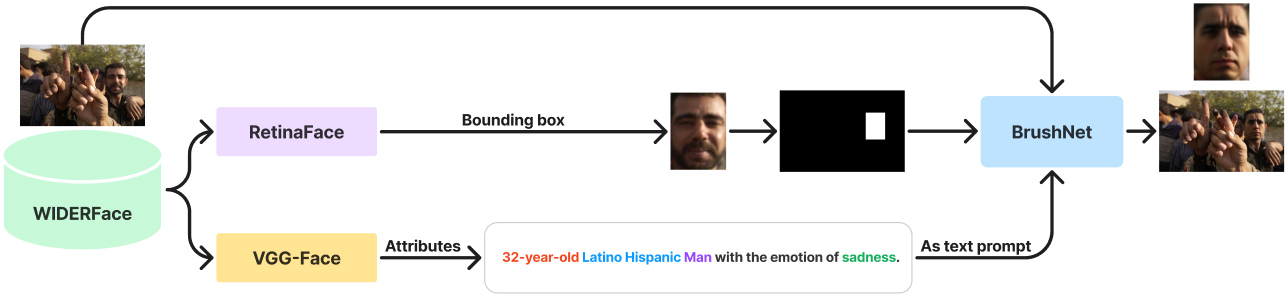}}
\caption{\textbf{Proposed pipeline.} Face detection is done by using RetinaFace. Facial attribute analysis is performed by VGG-Face and then these attributes are converted to text. Finally, faces are generated by BrushNet}
\label{fig:figure1}
\end{figure}

We introduce a three-stage pipeline for the face anonymization task. We first find the faces with RetinaFace and create binary masks for the face regions. Then, we extract age, gender, ethnicity, and facial expression attributes using VGG-Face.  
These extracted features are converted to text for the last stage. At the last stage, we feed BrushNet with the original image, face masks, and text prompt (\cite{brushnet}). The complete pipeline can be seen in Figure \ref{fig:figure1}.

We use RetinaFace in the face detection stage. RetinaFace achieves state-of-the-art results for face detection on the WIDER Face dataset (\cite{wider}). After we get face detections, we create binary masks for the detected facial areas.

We utilize VGG-Face to create text prompts that have facial attribute information. The VGG-Face model predicts the \textit{age}, \textit{ethnicity}, \textit{gender}, and \textit{expression} from the facial area. We then create texts from these predictions that are in a meaningful structure. One example text prompt is \textit{"32-year-old Latino Hispanic Man with the emotion of sadness."}

After we create masks and text prompts, we feed the BrushNet with them in addition to the original image. The model inpaint the original image in the mask area. We feed the model with each face mask in an image with the corresponding text prompt and generate a new face at each mask location.

\section*{Experiments}

\begin{figure}[htb]
\noindent
\centering
\includegraphics[width=0.1\textwidth, height=0.1\textwidth]{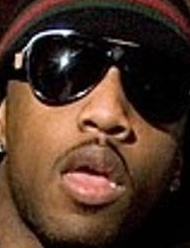}%
\includegraphics[width=0.1\textwidth, height=0.1\textwidth]{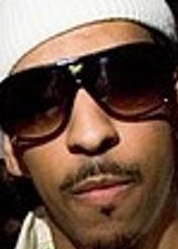}%
\includegraphics[width=0.1\textwidth, height=0.1\textwidth]{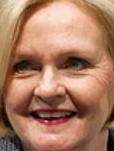}%
\includegraphics[width=0.1\textwidth, height=0.1\textwidth]{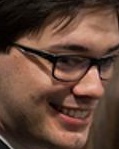}%
\includegraphics[width=0.1\textwidth, height=0.1\textwidth]{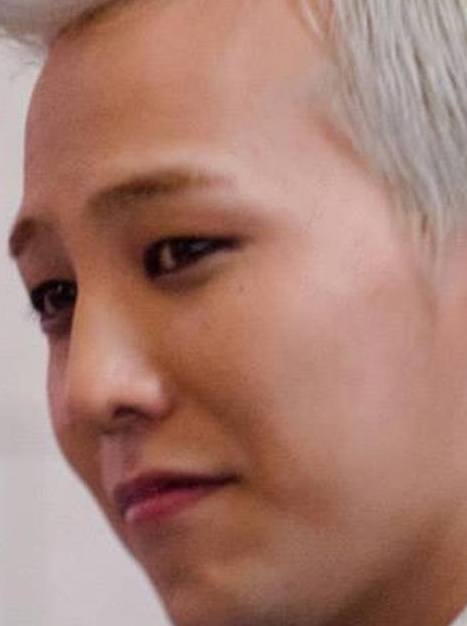}%
\includegraphics[width=0.1\textwidth, height=0.1\textwidth]{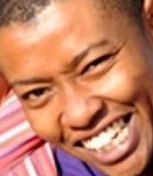}%
\includegraphics[width=0.1\textwidth, height=0.1\textwidth]{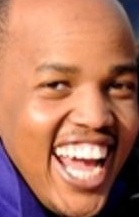}%
\includegraphics[width=0.1\textwidth, height=0.1\textwidth]{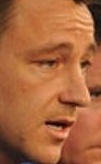}%
\includegraphics[width=0.1\textwidth, height=0.1\textwidth]{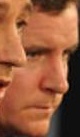}\\
\vspace{-1.00mm}
\includegraphics[width=0.1\textwidth, height=0.1\textwidth]{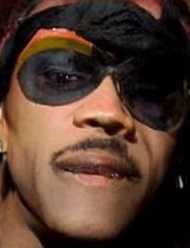}%
\includegraphics[width=0.1\textwidth, height=0.1\textwidth]{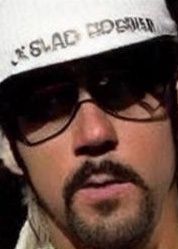}%
\includegraphics[width=0.1\textwidth, height=0.1\textwidth]{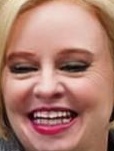}%
\includegraphics[width=0.1\textwidth, height=0.1\textwidth]{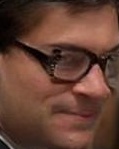}%
\includegraphics[width=0.1\textwidth, height=0.1\textwidth]{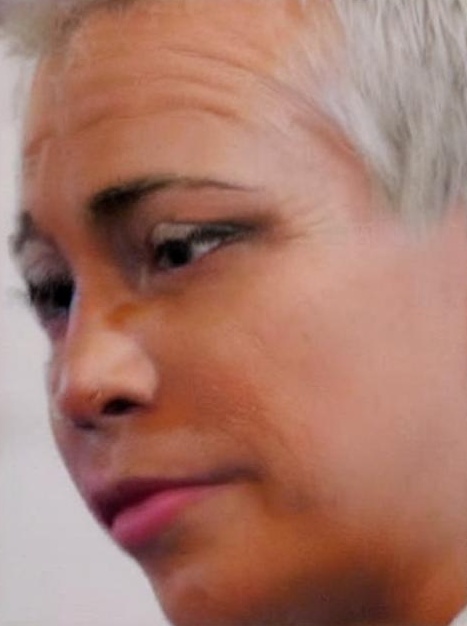}%
\includegraphics[width=0.1\textwidth, height=0.1\textwidth]{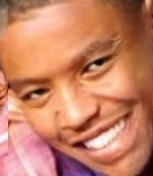}%
\includegraphics[width=0.1\textwidth, height=0.1\textwidth]{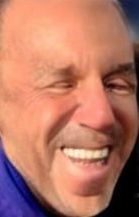}%
\includegraphics[width=0.1\textwidth, height=0.1\textwidth]{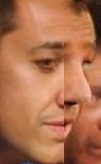}%
\includegraphics[width=0.1\textwidth, height=0.1\textwidth]{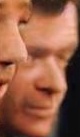}\\
\vspace{-1.00mm}
\includegraphics[width=0.1\textwidth, height=0.1\textwidth]{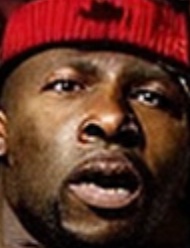}%
\includegraphics[width=0.1\textwidth, height=0.1\textwidth]{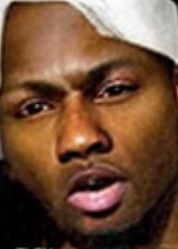}%
\includegraphics[width=0.1\textwidth, height=0.1\textwidth]{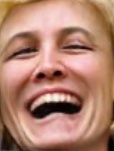}%
\includegraphics[width=0.1\textwidth, height=0.1\textwidth]{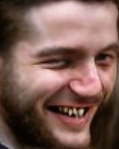}%
\includegraphics[width=0.1\textwidth, height=0.1\textwidth]{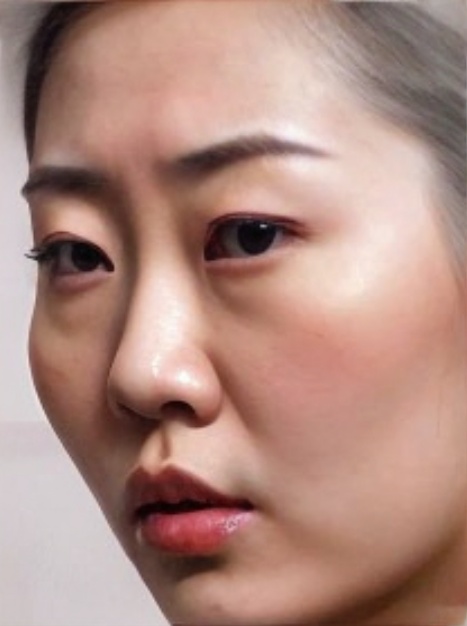}%
\includegraphics[width=0.1\textwidth, height=0.1\textwidth]{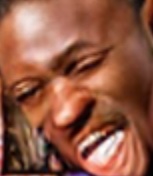}%
\includegraphics[width=0.1\textwidth, height=0.1\textwidth]{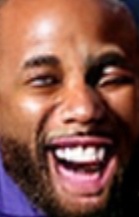}%
\includegraphics[width=0.1\textwidth, height=0.1\textwidth]{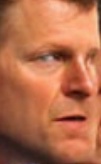}%
\includegraphics[width=0.1\textwidth, height=0.1\textwidth]{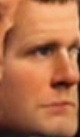}
\caption{\textbf{Qualitative results.} Example face anonymization results. The first row is the original images, the second row is the results of LDFA and the last row is results of our pipeline} 

\label{fig:figure2}\par
\end{figure}

We run the pipeline on WIDER Face validation data and provide qualitative results. Example faces generated by our pipeline can be seen in Figure \ref{fig:figure2}.

\begin{figure}[htb]
\noindent
\centering
\includegraphics[width=0.15\textwidth, height=0.15\textwidth]{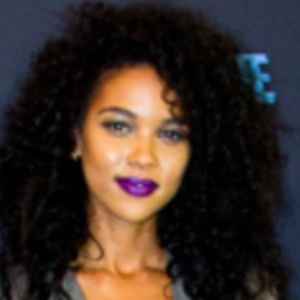}%
\includegraphics[width=0.15\textwidth, height=0.15\textwidth]{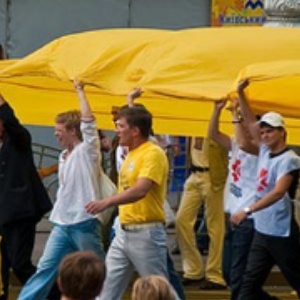}\\
\vspace{-1.00mm}
\includegraphics[width=0.15\textwidth, height=0.15\textwidth]{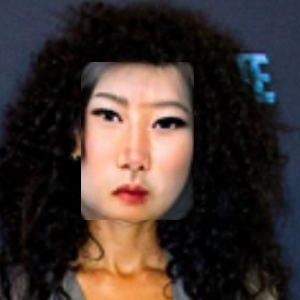}%
\includegraphics[width=0.15\textwidth, height=0.15\textwidth]{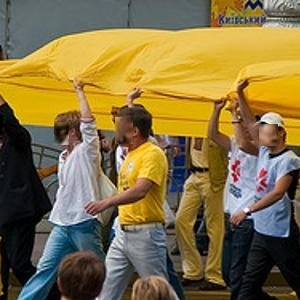}
\caption{\textbf{Limitations.} The top row is the original images and the bottom row is generated images. The first column is related to the capabilities of the inpainting model and the second one is mostly related to the resolution of the facial region}
\label{fig:figure3}\par
\end{figure}

From Figure \ref{fig:figure2}, we see that our pipeline can generate faces that are different from the original identity. However, it preserves the facial attributes. Since our pipeline consists of three different models, our method depends on the capabilities of these models. This dependency causes some limitations to our pipeline. In one example scenario, when the resolution of the face region extracted from RetinaFace is less than $100\times 100$ pixels, BrushNet struggles to generate face images. In another scenario, BrushNet may generate faces that do not integrate into the image seamlessly. These drawbacks of the pipeline can be seen in Figure \ref{fig:figure3}. 

\section*{Conclusions and Future Work}

In summary, with the growth of computer vision applications in our lives, computer vision researchers demand more image-based datasets than ever. Since the images collected in such datasets often include people's private information, researchers must ensure that data privacy regulations are employed when collecting images/videos. We introduce a new realistic face anonymization pipeline for this purpose. In our pipeline, we generate images that preserve the age, ethnicity, expression, and gender of the individuals by making use of facial attributes. Thus, by applying our pipeline to collected images for datasets, researchers can create natural-looking but unrecognizable people.

For future work, we plan to run quantitative experiments to analyze how much face recognition results change after image generation and how our pipeline affects facial attribute classification results. In addition, we plan to replace the rectangle-shaped masks with segmentation masks that are generated by a facial segmentation model in order to be able to integrate generated faces into the image more naturally.

\section*{Acknowledgements}

This work was partially supported by the European Union’s Horizon Europe research and innovation program under grant agreement No. 101135798 (My Personal AI Mediator for Virtual MEETtings BetWEEN People).





\normalsize

\small
\nocite{*}
\bibliographystyle{visuAAL-GoodBrother}
\bibliography{visuAAL-GoodBrother}

\normalsize

\end{document}